\pdfoutput=1

\documentclass[11pt]{article}

\usepackage{ACL2023}
\usepackage{times}
\usepackage{latexsym}
\usepackage{amsmath}
\usepackage[T1]{fontenc}

\usepackage[utf8]{inputenc}

\usepackage{microtype}

\usepackage{inconsolata}

\usepackage{xspace}
\usepackage{xcolor}
\usepackage{multirow}
\usepackage{booktabs}
\usepackage{tcolorbox}
\usepackage{caption}
\usepackage{dsfont}
\usepackage{float}
\usepackage{amssymb}
\usepackage{algorithm}
\usepackage{algorithmic}
\NewDocumentCommand\logo{}{
    \includegraphics[scale=0.25]{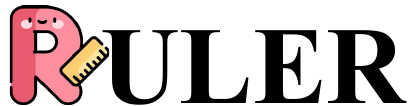}
}

\newcommand{\method}[0]{{\textsc{Ruler}}}

\title{\logo: A Model-Agnostic Method to Control Generated\\Length for Large Language Models}

\author{
Jiaming Li$^{1,2}$ \quad
Lei Zhang$^{1,2}$ \quad
Yunshui Li$^{1,2}$ \quad
Ziqiang Liu$^{1,2}$ \quad
Yuelin Bai$^{1,2}$ \quad
\\
\textbf{Run Luo}\textsuperscript{1,2} \quad
\textbf{Longze Chen}\textsuperscript{1,2}  \quad
\textbf{Min Yang}\textsuperscript{1}\footnotemark[2] \\
\textsuperscript{1}Shenzhen Institute of Advanced Technology, Chinese Academy of Sciences\\
\textsuperscript{2}University of Chinese Academy of Sciences\\
\texttt{\{jm.li4, min.yang\}@siat.ac.cn}
}

\definecolor{darkgreen}{rgb}{0.0, 0.5, 0.0}
\definecolor{darkred}{rgb}{0.55, 0.0, 0.0}

\begin{document}
\maketitle

\renewcommand{\thefootnote}{\fnsymbol{footnote}}
\footnotetext[2]{Min Yang is the corresponding author.}
\renewcommand{\thefootnote}{\arabic{footnote}}

\begin{abstract}

	The instruction-following ability of large language models enables humans to interact with AI agents in a natural way.
	However, when required to generate responses of a specific length, large language models often struggle to meet users' needs due to their inherent difficulty in accurately perceiving numerical constraints.
	To explore the ability of large language models to control the length of generated responses, we propose the Target Length Generation Task (\textit{TLG}) and design two metrics, Precise Match (PM) and Flexible Match (FM) to evaluate the model's performance in adhering to specified response lengths. Furthermore, we introduce a novel, model-agnostic approach called \method, which employs Meta Length Tokens (\textit{MLTs}) to enhance the instruction-following ability of large language models under length-constrained instructions.
	Specifically, \method{} equips LLMs with the ability to generate responses of a specified length based on length constraints within the instructions.
	Moreover, \method{} can automatically generate appropriate \textit{MLT} when length constraints are not explicitly provided, demonstrating excellent versatility and generalization.  Comprehensive experiments show the effectiveness of \method{} across different LLMs on Target Length Generation Task, e.g., at \textit{All Level} 27.97 average gain on PM, 29.57 average gain on FM.
	In addition, we conduct extensive ablation experiments to further substantiate the efficacy and generalization of \method. Our code and data is available at \url{https://github.com/Geaming2002/Ruler}.

\end{abstract}

\section{Introduction}

Large Language Models (LLMs) have demonstrated remarkable capabilities across a variety of natural language tasks and are increasingly being utilized in various fields \citep{vaswani2017attention,devlin-etal-2019-bert,brown2020language}. A primary area of interest is the instruction following ability, referring to their capability to execute tasks or generate outputs based on instructions \citep{ouyang2022training,wei2022finetuned}. It reflects the model's effectiveness in understanding and responding to instructions.

\begin{figure}[!t]
	\centering
	\includegraphics[width=\columnwidth]{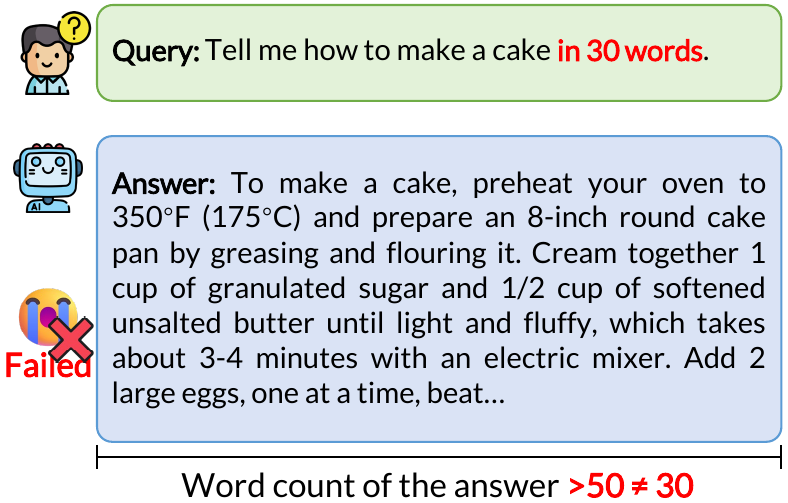}
	\caption{Existing LLMs lack the capability to follow instructions for generating texts of a specified length.}
	\label{fig:example}
\end{figure}

The practical challenges highlight the complexity of achieving precise instruction following, particularly when users require control over the output's length. Users frequently give LLMs various instructions, such as "Tell me how to make a cake in 20 words", "Use 50 words to write a post", "Write a 300-word story for me" and so on. These instructions challenge the instruction following capability of LLMs. To explore how well LLMs handle such challenges, we focus on the scenario where users specify the target length of the responses. The question is posed, "Can LLMs accurately generate with target length?" and introduce the \textit{Target Length Generation Task~(TLG)}. We create a test dataset with various target lengths and introduce two evaluation metrics: Precise Match (PM) and Flexible Match (FM). Our findings reveal that current LLMs generally perform poorly in this task, indicating considerable room for improvement. A discussion on the underlying causes is conducted, primarily attributing it to tokenization schemes and model training strategy.

To address aforementioned issues, we introduce \method, a model-agnostic approach designed to enhance the instruction-following capability of LLMs through \textit{Meta Length Tokens (MLTs)}. \textit{MLTs} are designed to control model's responses. By utilizing \method, LLMs can generate responses that meet target lengths. We create a dataset with \textit{MLTs} $\mathcal{D}_{MLT}$ for end-to-end training of LLMs. LLMs learn to generate \textit{MLT} and the corresponding length response after training. During inference, if a target length is provided, \method{} can transform it into a \textit{MLT} and generate responses that meet the requirement. If no target length is specified, it first generates a \textit{MLT}, then the response, ensuring its length aligns with the generated \textit{MLT}.

We apply \method{} to various large language models and test them on \textit{TLG}. Each model demonstrates significant improvements. Across all evaluated models, we observe a consistent improvement in both PM and FM scores at all \textit{Levels}. The PM and FM scores across \textit{All Level} showed an average improvement of 27.97 and 29.57.Furthermore, to rigorously test the capabilities of \method, we randomly sample the dataset provided by \citet{arenahard2024} and assess \method{} on multi \textit{MLT} generation and self-generated \textit{MLT} experiment to show the its effectiveness and generalizability. Additionally, \method{} is tested on six benchmarks to observe whether the models' overall performance is affected.

Our contributions can be summarized as follows:

\begin{itemize}
	\item We introduce the \textit{Target Length Generation Task (TLG)}, which designed to assess the instruction following capability of LLMs. It evaluates how well models generate responses of target lengths as directed by instructions.
	\item We propose \method, a novel and model-agnostic approach which employs the \textit{Meta Length Tokens (MLTs)}. Through end-to-end training, it enables models to generate response matching the target lengths indicated by \textit{MLTs}.
	\item We demonstrate that \method{} significantly enhances the performance of various models on the \textit{TLG}. Further experiments have also validated the effectiveness and generalizability of \method.
\end{itemize}

\begin{table*}[h]
	\centering
	\scalebox{0.9}{
		\begin{tabular}{lccc}
			\toprule
			\textbf{\textit{Level}}           & \textbf{Target Length} & \textbf{Precise Match (PM)} & \textbf{Flexible Match (FM)} \\
			\midrule
			\multirow{4}{*}{\textit{Level:0}} & 10                     & $ \pm 10$                   & $(0,20]$                     \\
			                                  & 30                     & $ \pm 10$                   & $(20,40]$                    \\
			                                  & 50                     & $ \pm 10$                   & $(40,60]$                    \\
			                                  & 80                     & $ \pm 10$                   & $(60,100]$                   \\
			\midrule
			\multirow{3}{*}{\textit{Level:1}} & 150                    & $ \pm 20$                   & $(100,200]$                  \\
			                                  & 300                    & $ \pm 20$                   & $(200,400]$                  \\
			                                  & 500                    & $ \pm 50$                   & $(400,600]$                  \\
			\midrule
			\multirow{2}{*}{\textit{Level:2}} & 700                    & $ \pm 70$                   & $(600,800]$                  \\
			                                  & >800                   & $(800,\infty)$              & $(800,\infty)$               \\
			\bottomrule
		\end{tabular}
	}
	\caption{\label{tab:target lengths}
		Nine target lengths and their corresponding match ranges categorized as Precise Match~(PM) and Flexible Match~(FM). Target lengths are classified into three categories, \textit{Level:0}, \textit{Level:1}, and \textit{Level:2}.
	}
\end{table*}

\section{Related Work}
\subsection{Large Language Model}

The advent of LLMs has revolutionized the field of natural language processing and become a milestone \citep{vaswani2017attention,devlin-etal-2019-bert,brown2020language, zhang2023marathon}. Large language models have achieved success across various NLP tasks. Models such as GPT-4\citep{achiam2023gpt}, Llama-3\citep{llama3modelcard}, and Qwen\citep{qwen}, known for their powerful capabilities, are increasingly serving as the foundation for various applications and making significant inroads into diverse fields, exerting a substantial impact. In-context learning enables LLMs to infer and generate responses solely based on the contextual information provided within a prompt\citep{dong2022survey,wei2022emergent}. This capability allows the models to exhibit a high degree of flexibility and adaptability across a variety of tasks\citep{levine2022the,chen-etal-2022-meta,zhao2021calibrate}. CoT further excavates and demonstrates the powerful logical reasoning capabilities of LLMs\citep{wei2022chain,huang-chang-2023-towards,zhang2023automatic}.

\subsection{Instruction Following}

Instruction following refers to the ability of large language models to comprehend and execute given natural language instructions \citep{brown2020language,ouyang2022training,wei2022finetuned,zhou2023lima}. This capability enables the models to perform a broad spectrum of tasks, from simple query responses to complex problem-solving and content generation, tailored to specific user requests.

In practical deployments, models may not adhere to comply with user instructions, exhibiting behaviors that deviate from anticipated outcomes. This includes generating responses unrelated to explicit instructions, emitting redundant or erroneous information, or entirely ignoring specified directives \citep{gehman-etal-2020-realtoxicityprompts,kenton2021alignment,wei2024jailbroken}. To enhance the instruction following capability of LLMs, open-domain instruction following data is frequently used for training. Several prominent studies have explored the construction of instruct-tuning data, to achieve efficient and cost-effective results\citep{li2024oneshot,cao2024instruction,liu2024what,xu2024wizardlm}.

\subsection{Meta Token}

Recently, an increasing number of studies have employed custom tokens within language models to execute specific functions or enhance performance. \citet{todd2024function} report findings that the hidden states of language models capture representations of these functions, which can be condensed into a Function Vector~(FV). Furthermore, their research demonstrates that FV can effectively guide language models in performing specific tasks.

Numerous studies have utilized meta tokens to compress prompts, thereby enhancing the inference capability of models \citep{li-etal-2023-compressing,liu-etal-2023-tcra,zhang2024soaring}. \citet{mu2023learning}introduce the concept of "gist tokens", which can be cached and reused for compute efficiency. Further \citet{jiang-etal-2024-hierarchical} utilize a hierarchical and dynamic approach to extend the concept, proposing "HD-Gist tokens" to improve model performance.

\section{Can LLMs Accurately Generate with Target Length?}
\label{sec:Can LLMs Accurately Generate with Target Length?}

In this section, we examine the capability of LLMs to generate responses of a target length. Initially, we introduce \textit{Target Length Generation Task~(TLG)}. Subsequently, we establish various target lengths and two evaluation metrics ($\S$\ref{subsec:Target Length Generation Task}). We then detail the experimental setup and assess the ability of LLMs to generate responses at target lengths ($\S$\ref{subsec:Experimental Setup}). Finally, we present the outcomes of the experiments and analysis the underlying reasons($\S$\ref{subsec:Results and Analysis}).

\begin{table*}[h]
	\centering
	\scalebox{0.9}{
		\begin{tabular}{lrrrrrrrrr}
			\toprule
			\multirow{3}{*}{\textbf{Model}} & \multirow{3}{*}{\textbf{Params}} & \multicolumn{8}{c}{\textit{Target Length Generation Task~(TLG)}}                                                                                                                                                                                                        \\
			\cmidrule(lr){3-10}
			                                &                                  & \multicolumn{2}{c}{\textit{Level:0}}                             & \multicolumn{2}{c}{\textit{Level:1}} & \multicolumn{2}{c}{\textit{Level:2}} & \multicolumn{2}{c}{\textit{All Level}}                                                                                 \\
			\cmidrule(lr){3-4}\cmidrule(lr){5-6}\cmidrule(lr){7-8}\cmidrule(lr){9-10}
			                                &                                  & PM                                                               & FM                                   & PM                                   & FM                                     & PM                & FM                & PM                & FM                \\
			\midrule
			\multicolumn{10}{c}{\textit{Closed-source Model}\footnotemark[1]}                                                                                                                                                                                                                                                                            \\
			\midrule
			gpt-4-turbo                     & -                                & \textbf{82.26}                                                   & \textbf{86.36}                       & \textbf{46.49}                       & \textbf{85.06}                         & 40.72             & 47.51             & \underline{61.35} & \underline{77.35} \\
			gpt-4o                          & -                                & 74.06                                                            & 79.05                                & 32.32                                & 69.36                                  & 36.22             & \textbf{71.95}    & 57.75             & 74.30             \\
			gpt-3.5-turbo                   & -                                & 64.41                                                            & 69.84                                & 35.06                                & 75.76                                  & 38.24             & 45.93             & 49.00             & 66.50             \\
			\midrule
			claude-3-haiku                  & -                                & 48.23                                                            & 55.21                                & 35.37                                & 73.78                                  & \underline{44.12} & 50.45             & 43.10             & 60.25             \\
			claude-3.5-sonnet               & -                                & \underline{75.17}                                                & \underline{81.04}                    & \underline{42.38}                    & \underline{83.08}                      & \textbf{62.67}    & \underline{71.27} & \textbf{61.65}    & \textbf{79.55}    \\
			\midrule
			\multicolumn{10}{c}{\textit{Open-source Model}}                                                                                                                                                                                                                                                                                              \\
			\midrule
			Mistral                         & 7B                               & 20.29                                                            & 23.50                                & 16.77                                & 48.32                                  & 3.62              & 5.66              & 15.45             & 27.70             \\
			\midrule
			\multirow{2}{*}{Gemma}          & 2B                               & 20.95                                                            & 23.17                                & 8.69                                 & 24.24                                  & 0.23              & 0.23              & 12.35             & 18.45             \\
			                                & 7B                               & 15.52                                                            & 18.85                                & 11.74                                & 35.82                                  & 0.45              & 0.45              & 10.95             & 20.35             \\
			\midrule
			\multirow{2}{*}{Llama3}         & 8B                               & 34.59                                                            & \underline{40.02}                    & \underline{29.73}                    & \underline{65.70}                      & 18.10             & 21.04             & \underline{29.35} & \underline{44.25} \\
			                                & 70B                              & \textbf{58.76}                                                   & \textbf{64.52}                       & \textbf{36.59}                       & \textbf{77.90}                         & \textbf{36.43}    & \textbf{41.18}    & \textbf{46.55}    & \textbf{63.75}    \\
			\midrule
			\multirow{2}{*}{InternLM2}      & 7B                               & 6.65                                                             & 7.21                                 & 8.69                                 & 27.44                                  & 19.68             & 22.40             & 10.20             & 17.20             \\
			                                & 20B                              & 8.98                                                             & 9.87                                 & 10.98                                & 34.45                                  & 17.42             & 20.14             & 11.50             & 20.20             \\
			\midrule
			\multirow{2}{*}{DeepSeek-LLM}   & 7B                               & 28.16                                                            & 31.37                                & 17.68                                & 44.36                                  & 10.86             & 13.12             & 20.90             & 31.60             \\
			                                & 67B                              & 26.94                                                            & 30.27                                & 17.07                                & 49.54                                  & 9.50              & 11.99             & 19.85             & 32.55             \\
			\midrule
			\multirow{3}{*}{Yi-1.5}         & 6B                               & 23.50                                                            & 25.83                                & 16.46                                & 48.78                                  & 18.10             & 20.36             & 20.00             & 32.15             \\
			                                & 9B                               & 25.28                                                            & 29.16                                & 17.38                                & 44.36                                  & \underline{24.43} & \underline{29.41} & 22.50             & 34.20             \\
			                                & 34B                              & 28.82                                                            & 33.59                                & 26.07                                & 65.40                                  & 21.27             & 25.79             & 26.25             & 42.30             \\
			\midrule
			\multirow{4}{*}{Qwen1.5}        & 7B                               & 24.28                                                            & 27.38                                & 14.33                                & 46.19                                  & 9.05              & 11.99             & 17.65             & 30.15             \\
			                                & 14B                              & 28.27                                                            & 31.49                                & 18.45                                & 43.90                                  & 11.09             & 14.25             & 21.25             & 31.75             \\
			                                & 32B                              & 32.59                                                            & 36.25                                & 22.26                                & 49.39                                  & 21.49             & 25.34             & 26.75             & 38.15             \\
			                                & 72B                              & \underline{35.59}                                                & 39.69                                & 18.29                                & 49.70                                  & 3.85              & 6.11              & 22.90             & 35.55             \\
			\bottomrule
		\end{tabular}
	}
	\caption{\label{tab:TLG Results}
		Overall results of different LLMs of \textit{TLG}. All open-source models used are either chat or instruct models. In models belonging to the same series but varying in parameter sizes, those with larger parameters typically exhibit superior performance. The best-performing model in each \textit{Level} is \textbf{in-bold}, and the second best is \underline{underlined}.
	}
\end{table*}

\subsection{Target Length Generation Task}
\label{subsec:Target Length Generation Task}

To assess the ability of existing LLMs to control the length of generated response, we develop the \textit{TLG}. This task assesses the models' ability in producing responses that match target lengths as directed designed target lengths are detailed in Table \ref{tab:target lengths}. Additionally, we divide these nine target lengths into three \textit{levels}: \textit{Level:0}, \textit{Level:1}, and \textit{Level:2}.

Given that generating responses with target lengths is challenging for existing LLMs, we develop two metrics to evaluate the accuracy of response lengths.

\begin{itemize}
	\item \textbf{Precise Match (PM):} This metric requires that the length of the generated response be very close to the target length. For different \textit{Level}, a precise tolerance range is set ($\pm 10$, $\pm 20$, \dots) necessitating that the response length stringently conforms to these defined limits.
	\item \textbf{Flexible Match (FM):} This metric requires a broader tolerance interval for target length. For longer texts, the range incrementally widens to meet response generation requirements.
\end{itemize}

For the $N$ responses, we assess whether response meets the target length, then calculating the PM and FM scores of the model.

\begin{equation}
	\label{eq:PM}
	\mathrm{PM}=\frac{\sum_{i=1}^{N}\mathds{1}\left(\mathrm{lb_{TL_i}^{P}}<L\left(c_i\right)\leq \mathrm{ub_{TL_i}^{P}}\right)}{N}
\end{equation}

\begin{equation}
	\label{eq:FM}
	\mathrm{FM}=\frac{\sum_{i=1}^{N}\mathds{1}\left(\mathrm{lb_{TL_i}^{F}}<L\left(c_i\right)\leq \mathrm{ub_{TL_i}^{F}}\right)}{N}
\end{equation}
where: $c_i$ denotes the $i$-th response generated by LLM. The function $L(\cdot)$ calculates the word count of the input string. $\mathrm{lb_{TL_i}^{P}}$ and $\mathrm{ub_{TL_i}^{P}}$ denote the lower and upper bounds of the precise match range associated with the target length of $i$-th response. $\mathrm{lb_{TL_i}^{F}}$ and $\mathrm{ub_{TL_i}^{F}}$ denote the lower and upper bounds of the flexible match range associated with the target length of $i$-th response.

\subsection{Experimental Setup}
\label{subsec:Experimental Setup}

\paragraph{Dataset.}

We employ a two-stage data construction method for this study. Initially, we randomly sample 2,000 data from OpenHermes2.5 \citep{OpenHermes2.5}. To enhance the complexity of the task and prevent data leakage, the second stage involved uses only the questions from these samples. Additionally, we randomly assign one of nine target lengths for the responses. The distribution of target length in the \textit{TLG} dataset is shown in Figure \ref{fig:TLG}. Further details regarding the format of the \textit{TLG} dataset are provided in Appendix \ref{appendix:Dataset of TLG}.

\begin{figure*}[!htbp]
	\centering
	\includegraphics[width=\textwidth]{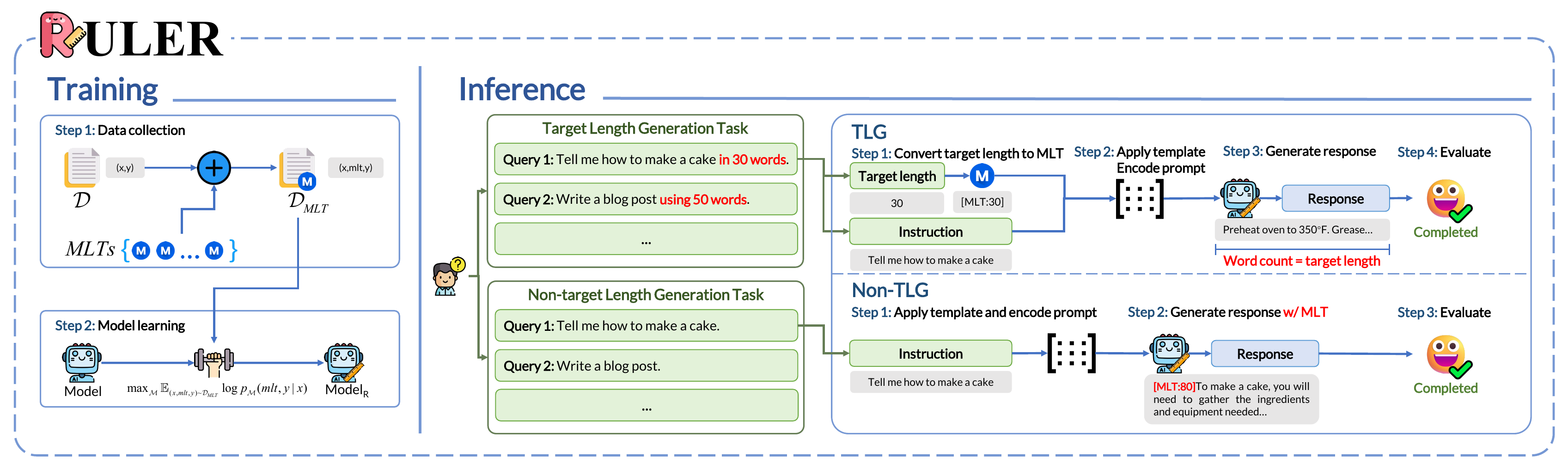}
	\caption{Overview of \method. The method is divided into two parts: training and inference. The figure illustrates the main content of both sections. Additionally, in the inference section, we show two scenarios: \textit{TLG} and \textit{non-TLG} to show the difference.
	}
	\vspace{-3mm}
	\label{fig:method}
\end{figure*}

\begin{figure}[!t]
	\centering
	\includegraphics[width=0.7\columnwidth]{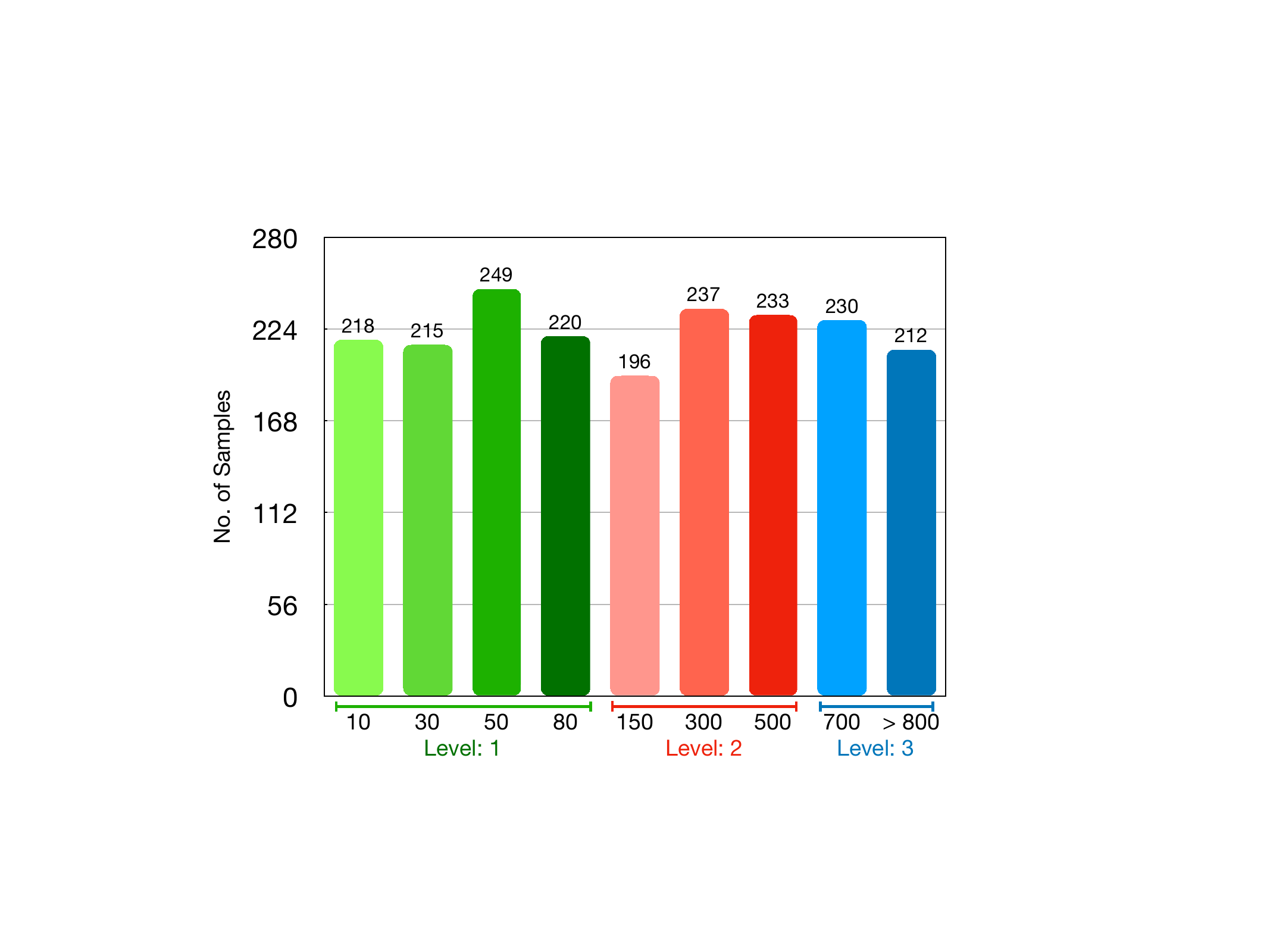}
	\caption{Target length distribution in \textit{TLG} dataset. The count of each target length is approximately 200.}
	\label{fig:TLG}
\end{figure}

\footnotetext[1]{The results of all closed-source models are obtained on July 26, 2024.}

\paragraph{Models \& Prompt Templates.}

We conduct extensive experiments with both closed and open-source LLMs, specifically the chat or instruct version. The specific models used are listed in Table \ref{tab:models}. We evaluate each model using its own prompt template, as detailed in Table \ref{tab:Models-templates}.

To integrate the target length into the prompt, we modify the sentence \texttt{The response should have a word count of \{Target Length\} words} into each question. For target length >800, we replace this with \texttt{more than 800}.

\paragraph{Hardware \& Hyperparameters.}

All experiments are conducted on NVIDIA A100 GPUs. Inference is performed using the \texttt{vllm} \citep{kwon2023efficient}, with \texttt{temperature} set to 0 and \texttt{max\_tokens} set to 2,048 in the \texttt{SamplingParams}, thereby employing greedy decoding for inference. The \texttt{model\_max\_length} for all models is consistent with their respective  configurations, as shown in Table \ref{tab:models}.

\subsection{Results and Analysis}
\label{subsec:Results and Analysis}

Table \ref{tab:TLG Results} displays the PM and FM scores of open-source models at different \textit{Levels}. Generally, models with advanced  capabilities achieve higher PM and FM scores, indicating stronger adherence to instructions. This observation aligns with human expectations.

For most models, scores are lowest at \textit{Level:2}, suggesting significant potential for enhancement in producing longer responses. While, scores at \textit{Level:1} are the highest. This trend may be attributed to the prevalence of shorter responses in the training datasets utilized for model fine-tuning, which influences their generative biases. Desipte potential differences in parameters, a performance gap between closed and open source models remains evident. Notably, claude-3.5-Sonnet achieve the best scores across all models at the \textit{All Level}, with scores of 61.65 and 79.55. Furthermore, the PM and FM scores for each model across various target lengths are detailed in Appendix \ref{appendix:Results on Different Target Length}.

The poor performance in \textit{TLG} can be attributed to a discrepancy between the token counts generated by LLMs and the lengths as understood by humans. The discrepancy between the tokens generated by LLMs and the lengths as understood by humans constirbutes to the issue. This mismatch arises due to several factors:

\begin{itemize}
	\item \textbf{Tokenization Schemes:} LLMs employ subword tokenization schemes that decompose words into smaller units of varying lengths. For example, a single long word might be divided into multiple tokens, complicating the model's ability to equate token counts with human-understood word counts \citep{Gage1994ANA}.
	\item \textbf{Model Training:} Most LLMs, particularly those trained using autoregressive language modeling, are not explicitly trained with objectives that prioritize output length. As a result, these models often lack strong capabilities for controlling the length of their generated output\citep{devlin-etal-2019-bert}.
\end{itemize}

\begin{table}[!t]
	\centering
	\scalebox{0.8}{
		\begin{tabular}{l|c|c}
			\toprule
			\textbf{\textit{MLT}} & \textbf{Range of Variation} & \textbf{No. in $\mathcal{D}_{MLT}$} \\
			\midrule
			\texttt{[MLT:10]}     & $[5,15)$                    & 20,000                              \\
			\texttt{[MLT:30]}     & $[25,35)$                   & 20,000                              \\
			\texttt{[MLT:50]}     & $[45,55)$                   & 20,000                              \\
			\texttt{[MLT:80]}     & $[75,85)$                   & 20,000                              \\
			\midrule
			\texttt{[MLT:150]}    & $[145,155)$                 & 20,000                              \\
			\texttt{[MLT:300]}    & $[295,305)$                 & 10,333                              \\
			\texttt{[MLT:500]}    & $[495,505)$                 & 2,317                               \\
			\midrule
			\texttt{[MLT:700]}    & $[695,705)$                 & 497                                 \\
			\texttt{[MLT:>800]}   & $(800,\infty)$              & 8,082                               \\
			\bottomrule
		\end{tabular}
	}
	\caption{\label{table:MLT}
		Meta length tokens in \method{} showing their range of variation in data collection and counts in $\mathcal{D}_{MLT}$.
	}
\end{table}

\section{\method: Meta Length Token Controlled Generation}

In this section, we first introduce \method, encompassing the design of the \textit{Meta Length Tokens (MLTs)}, the data collection and the learning process associated with the models ($\S$\ref{subsec:Method}). Subsequently, we detail the difference in the generation of \method{} under two scenarios: \textit{TLG} and non-\textit{TLG} ($\S$\ref{subsec:method Inference}).

\subsection{Method}
\label{subsec:Method}

\paragraph{\method.}

We introduce \method, as illustrated in Figure \ref{fig:method}, to effectively control the response length of LLMs using \textit{MLTs}. Ruler employs MLTs to explicitly communicate length requirements within instructions. The \textit{MLTs} represent the model's response length range and aim to enhance its capability on the \textit{TLG} task. Our end-to-end training enables the LLMs to automatically generate \textit{MLTs} in various scenarios, regardless of target length requirements. \textit{MLTs} (Table \ref{table:MLT}) offer more precise control than traditional text prompt methods, which often prove insufficiently constraining.

\begin{table*}[h]
	\centering
	\scalebox{0.75}{
		\begin{tabular}{lllllllll}
			\toprule
			\multirow{3}{*}{\textbf{Model}} & \multicolumn{8}{c}{\textit{Target Length Generation Task (TLG)}}                                                                                                                                                                                                                                                                                                                                                                             \\
			\cmidrule(lr){2-9}
			                                & \multicolumn{2}{c}{\textit{Level:0}}                             & \multicolumn{2}{c}{\textit{Level:1}}              & \multicolumn{2}{c}{\textit{Level:2}}              & \multicolumn{2}{c}{\textit{All Level}}                                                                                                                                                                                                                            \\
			\cmidrule(lr){2-3}\cmidrule(lr){4-5}\cmidrule(lr){6-7}\cmidrule(lr){8-9}
			                                & PM                                                               & FM                                                & PM                                                & FM                                                & PM                                                & FM                                                & PM                                                & FM                                                \\
			\midrule
			Mistral-7B-Instruct             & 20.29                                                            & 23.50                                             & 16.77                                             & 48.32                                             & 3.62                                              & 5.66                                              & 15.45                                             & 27.70                                             \\
			Mistral-7B$_\textsc{R}$         & 70.18\small\textcolor{darkgreen}{$\uparrow$49.89}                & 75.06\small\textcolor{darkgreen}{$\uparrow$51.56} & 35.52\small\textcolor{darkgreen}{$\uparrow$18.75} & 67.84\small\textcolor{darkgreen}{$\uparrow$19.52} & 33.71\small\textcolor{darkgreen}{$\uparrow$30.09} & 36.43\small\textcolor{darkgreen}{$\uparrow$30.77} & 50.75\small\textcolor{darkgreen}{$\uparrow$35.30} & 64.15\small\textcolor{darkgreen}{$\uparrow$36.45} \\
			\midrule
			gemma-7b-it                     & 15.52                                                            & 18.85                                             & 11.74                                             & 35.82                                             & 0.45                                              & 0.45                                              & 10.95                                             & 20.35                                             \\
			gemma-7b$_\textsc{R}$           & 59.53\small\textcolor{darkgreen}{$\uparrow$44.01}                & 64.19\small\textcolor{darkgreen}{$\uparrow$45.34} & 39.33\small\textcolor{darkgreen}{$\uparrow$27.59} & 68.14\small\textcolor{darkgreen}{$\uparrow$32.32} & 25.34\small\textcolor{darkgreen}{$\uparrow$24.89} & 27.83\small\textcolor{darkgreen}{$\uparrow$27.38} & 45.35\small\textcolor{darkgreen}{$\uparrow$34.40} & 57.45\small\textcolor{darkgreen}{$\uparrow$37.10} \\
			\midrule
			Llama-3-8B-Instruct             & 34.59                                                            & 40.02                                             & 29.73                                             & 65.70                                             & 18.10                                             & 21.04                                             & 29.35                                             & 44.25                                             \\
			Llama-3-8B$_\textsc{R}$         & 77.27\small\textcolor{darkgreen}{$\uparrow$42.68}                & 80.71\small\textcolor{darkgreen}{$\uparrow$40.69} & 50.76\small\textcolor{darkgreen}{$\uparrow$21.03} & 83.84\small\textcolor{darkgreen}{$\uparrow$18.14} & 19.23\small\textcolor{darkgreen}{$\uparrow$1.13}  & 22.85\small\textcolor{darkgreen}{$\uparrow$1.81}  & 55.75\small\textcolor{darkgreen}{$\uparrow$26.40} & 68.95\small\textcolor{darkgreen}{$\uparrow$24.70} \\
			\midrule
			deepseek-llm-7b-chat            & 28.16                                                            & 31.37                                             & 17.68                                             & 44.36                                             & 10.86                                             & 13.12                                             & 20.90                                             & 31.60                                             \\
			deepseek-llm-7b$_\textsc{R}$    & 68.18\small\textcolor{darkgreen}{$\uparrow$40.02}                & 73.50\small\textcolor{darkgreen}{$\uparrow$42.13} & 31.10\small\textcolor{darkgreen}{$\uparrow$13.42} & 68.90\small\textcolor{darkgreen}{$\uparrow$24.54} & 11.54\small\textcolor{darkgreen}{$\uparrow$0.68}  & 11.76\small\textcolor{darkred}{$\downarrow$-1.36} & 43.50\small\textcolor{darkgreen}{$\uparrow$22.60} & 58.35\small\textcolor{darkgreen}{$\uparrow$26.75} \\
			\midrule
			Yi-1.5-6B-Chat                  & 23.50                                                            & 25.83                                             & 16.46                                             & 48.78                                             & 18.10                                             & 20.36                                             & 20.00                                             & 32.15                                             \\
			Yi-1.5-6B$_\textsc{R}$          & 67.07\small\textcolor{darkgreen}{$\uparrow$43.57}                & 72.17\small\textcolor{darkgreen}{$\uparrow$46.34} & 40.40\small\textcolor{darkgreen}{$\uparrow$23.94} & 76.83\small\textcolor{darkgreen}{$\uparrow$28.05} & 19.23\small\textcolor{darkgreen}{$\uparrow$1.13}  & 21.04\small\textcolor{darkgreen}{$\uparrow$0.68}  & 47.75\small\textcolor{darkgreen}{$\uparrow$27.75} & 62.40\small\textcolor{darkgreen}{$\uparrow$30.25} \\
			\midrule
			Qwen1.5-7B-Chat                 & 24.28                                                            & 27.38                                             & 14.33                                             & 46.19                                             & 9.05                                              & 11.99                                             & 17.65                                             & 30.15                                             \\
			Qwen1.5-7B$_\textsc{R}$         & 59.09\small\textcolor{darkgreen}{$\uparrow$34.81}                & 64.41\small\textcolor{darkgreen}{$\uparrow$37.03} & 29.88\small\textcolor{darkgreen}{$\uparrow$15.55} & 61.28\small\textcolor{darkgreen}{$\uparrow$15.09} & 11.54\small\textcolor{darkgreen}{$\uparrow$2.49}  & 14.25\small\textcolor{darkgreen}{$\uparrow$2.26}  & 39.00\small\textcolor{darkgreen}{$\uparrow$21.35} & 52.30\small\textcolor{darkgreen}{$\uparrow$22.15} \\
			\bottomrule
		\end{tabular}
	}

	\caption{\label{tab:MLT TLG results}
		Overall results of various LLMs with \method{} are presented. Additionally, we also annotate the table with the score changes compared to the chat or instruct model. Consistent improvements in both PM and FM scores are observed across all \text{Levels}.
	}
\end{table*}

\paragraph{Data collection for \method.}

For common fine-tuning training datasets, the format typically consist of input-output pairs $(x,y)$. Following \citet{zhou2023instruction}, we calculate the word count of $y$ for each entry. Based on the predefined \textit{MLTs} in Table \ref{table:MLT} and their range of variation, we aim to match each $y$ to a corresponding \textit{mlt} based on its word count. If a match is found, the data is reformatted as $(x,mlt,y)$. This method aids in the construction of the fine-tuning training dataset $\mathcal{D}_{MLT}$, detailed in Algorithm \ref{appendix:Data Creation}.

\paragraph{\method{} learning.}

To minimize changes to the model's generation pattern and ensure stability in non-\textit{TLG} scenario, we position the \textit{MLT} immediately before the original response during the construction of fine-tuning data. This strategy maintains the model chat template. Consequently, the combination of $mlt$ and the original response $y$ forms a new complete response $y^\prime$.

We conduct the training of the \method{} $\mathcal{M}$ on the curated corpus $\mathcal{D}_{MLT}$, which is augmented with \textit{Meta Length Tokens} $\mathcal{D}_{MLT}$, employing the standard next token objective:

\begin{equation}
	\label{eq:learning}
	\max_{\mathcal{M}}\mathbb{E}_{(x,mlt,y)\thicksim\mathcal{D}_{MLT}}\log p_{\mathcal{M}}(mlt,y|x)
\end{equation}

We concatenate the \textit{MLT} directly to the beginning of $y$ to compute the loss and use the \textit{MLTs} to expand the original vocabulary $\mathcal{V}$.

\subsection{\method{} Inference}
\label{subsec:method Inference}

\paragraph{\textit{TLG} scenario.}

In the \textit{Target Length Generation (TLG)} scenario, the user's instruction specifies a target length, decomposed into a question and a target length. The \method{} converts this target length into the corresponding \textit{MLT} and appends it to the model chat template. Subsequent to the \textit{MLT}, \method{} generates response that aligns with the target length, ensuring compliance with both the user's question and the target length, as illustrated in Figure \ref{fig:method}. This approach yields superior results compared to controlling outputs solely through prompts.

\paragraph{non-\textit{TLG} scenario.}

In the non-\textit{TLG} scenario, users provide straightforward instructions consisting solely of a question. \method{} integrates these instructions directly into the model's chat template for generation. Owing to its innovative design and the use of a standard next-token objective in training (Equation \ref{eq:learning}), \method{} autonomously generates a \textit{MLT} prior to producing the textual response. This \textit{MLT} is designed to match the length of the content generated, thereby ensuring normal generation of the model in non-\textit{TLG} scenarios, as illustrated in Figure \ref{fig:method}.

\section{Experiments}

\subsection{Experimental Setup}

\paragraph{Dataset $\mathcal{D}_{MLT}$.}

To ensure balanced frequency distribution of each \textit{Meta Length Token (MLT)} in $\mathcal{D}_{MLT}$, we set a maximum occurrence limit of 20,000 for each \textit{MLT}. We construct $\mathcal{D}_{MLT}$ from three datasets: OpenHermes2.5 (excluding data previously used in \textit{TLG}) \citep{OpenHermes2.5}, LongForm \citep{koksal2023longform}, and ELI5 \citep{fan2019eli5}, in accordance with Algorithm \ref{algorithm1}. This approach aims to create a diverse dataset, particularly effective for generating longer content that is relatively rare. in total, $\mathcal{D}_{MLT}$ comprises 121,229 entries, with the frequency of each \textit{MLT} in Table \ref{table:MLT}. Moreover, we calculate the word count for each response in every dataset, allowing us to statistically analyze the \textit{MLT} distribution, as detailed in Table \ref{tab:MLT in Datasets}.

\paragraph{LLMs.}

To comprehensively evaluate the performance of \method{} across different models, we consider factors such as model size, open-source availability, and overall model performance. We select six LLMs are selected: Mistral-7B-v0.3 \citep{jiang2023mistral}, gemma-7b \citep{gemmateam2024gemma}, Llama-3-8B \citep{llama3modelcard}, deepseek-llm-7b \citep{deepseek-llm}, Yi-1.5-6B \citep{ai2024yi}, and Qwen1.5-7B \citep{qwen}. We apply the \method{} to these base models and compare the results with their corresponding instruct or chat models.

\paragraph{Evaluation Metric.}

Consistent with the \textit{TLG} and compared to previous results, we also calculate PM and FM scores to assess the effectiveness of \method.

\subsection{Main Results}

Table \ref{tab:MLT TLG results} presents a detailed comparison of PM and FM scores across various LLMs using \method{} across different \textit{Levels}. For information on model training see Appendix \ref{appendix:More Details of Training}.

\paragraph{Overall Performance Enhancement.}

Across all evaluated models, we observe a consistent improvement in both PM and FM scores at all \textit{Levels}. The most significant improvement is observed in gemma-7b$_{\textsc{R}}$\footnote{Model name with $_{\textsc{R}}$ means base model with \method{}}, with PM and FM scores increasing by 34.40 and 37.10, respectively. In contrast, the least improvement is noted with PM and FM rising by 21.35 and 22.15. The PM and FM scores across \textit{All Level} showed an average improvement of 27.97 and 29.57. These improvements indicate that \method{} effectively enhances the model's ability to generate content of target lengths. This suggests that using \textit{MLT} to control output length is more effective than using prompts, as the model learns to generate content of corresponding lengths during fine-tuning. Additionally, \method's ability to enhance various models demonstrates its generalizability and scalability.

\paragraph{Different \textit{Level} Analysis.}

At \textit{Level:0}, all models show significant improvements in both PM and FM scores. Compared to other \textit{Level}, each model achieves the highest PM and FM score improvements at \textit{Level:0}. This enhancement occurs because the models are capable of generating responses of this length; however, their coarse length control impedes precise adherence to target length requirements. Our method significantly improves the models' capacity to accurately control content length at \textit{Level:0} more accurately, better meeting the target length requirements.

Moving to \textit{Level:1}, while the improvements are not as pronounced as at \textit{Level:0}, the models still exhibit significant gains in both PM and FM scores. At \textit{Level:2}, the extent of score improvements varies across models. For instance, Mistral-7B-v0.3$_{\textsc{R}}$ and gemma-7b$_{\textsc{R}}$ continue to show substantial score increases. Despite these positive trends, only deepseek-llm-7b-chat$_{\textsc{R}}$, show a slight decrease in scores at \textit{Level:2}. This is attributed to the insufficient data for \textit{Level:2} in $\mathcal{D}_{MLT}$. The uneven distribution of data likely contributes to the slight decrease in scores.

\begin{table*}[h]
	\centering
	\scalebox{0.9}{
		\begin{tabular}{l|rrrrrrrrr|c}
			\toprule
			\multirow{2}{*}{\textbf{Model}} & \multicolumn{9}{|c|}{FM of Different Target Length} & \multirow{2}{*}{\textbf{Avg FM}}                                                          \\
			\cmidrule(lr){2-10}
			                                & 10                                                  & 30                               & 50   & 80   & 150  & 300  & 500  & 700  & >800         \\
			\midrule
			Mistral-7B-Instruct-v0.3        & 0.5                                                 & 0.0                              & 0.5  & 2.0  & 18.5 & 50.5 & 20.5 & 3.0  & 2.5  & 10.89 \\
			Mistral-7B-v0.3$_{\textsc{R}}$  & 72.5                                                & 68.0                             & 65.5 & 76.5 & 76.0 & 63.0 & 28.0 & 24.0 & 64.5 & 59.78 \\
			\midrule
			gemma-7b-it                     & 13.0                                                & 17.0                             & 15.5 & 26.0 & 54.5 & 76.5 & 17.5 & 0.0  & 0.0  & 24.44 \\
			gemma-7b$_{\textsc{R}}$         & 58.0                                                & 63.5                             & 61.0 & 69.5 & 72.5 & 64.0 & 42.0 & 17.0 & 67.0 & 57.17 \\
			\midrule
			Llama-3-8B-Instruct             & 23.5                                                & 18.0                             & 12.5 & 28.0 & 50.5 & 76.5 & 57.0 & 25.5 & 30.5 & 35.78 \\
			Llama-3-8B$_{\textsc{R}}$       & 84.0                                                & 84.0                             & 73.0 & 80.0 & 87.5 & 89.5 & 71.0 & 14.5 & 36.5 & 68.89 \\
			\midrule
			deepseek-llm-7b-chat            & 36.5                                                & 16.0                             & 12.5 & 17.5 & 23.5 & 60.5 & 36.5 & 16.0 & 22.5 & 26.83 \\
			deepseek-llm-7b$_{\textsc{R}}$  & 64.0                                                & 70.0                             & 62.5 & 73.0 & 82.0 & 86.5 & 27.0 & 17.0 & 40.5 & 58.06 \\
			\midrule
			Yi-1.5-6B-Chat                  & 26.5                                                & 16.5                             & 14.5 & 14.5 & 18.5 & 42.5 & 35.0 & 33.5 & 28.5 & 25.56 \\
			Yi-1.5-6B$_{\textsc{R}}$        & 80.5                                                & 66.0                             & 67.0 & 77.0 & 83.5 & 83.5 & 56.0 & 22.0 & 39.5 & 63.89 \\
			\midrule
			Qwen1.5-7B-Chat                 & 13.5                                                & 17.0                             & 9.5  & 16.0 & 6.5  & 51.0 & 57.5 & 22.5 & 4.5  & 22.00 \\
			Qwen1.5-7B$_{\textsc{R}}$       & 69.0                                                & 61.0                             & 46.5 & 68.5 & 81.0 & 80.5 & 38.5 & 16.5 & 36.5 & 55.33 \\
			\bottomrule
		\end{tabular}
	}
	\caption{\label{tab:arena tl results}
		Results in multi \textit{MLT} generation experiment. Generally, the FM scores obtained via \method{} surpass those of the baseline models.
	}
\end{table*}

\begin{figure*}[htbp]
	\centering
	\begin{minipage}[c]{0.5\textwidth}
		\centering
		\includegraphics[width=\textwidth]{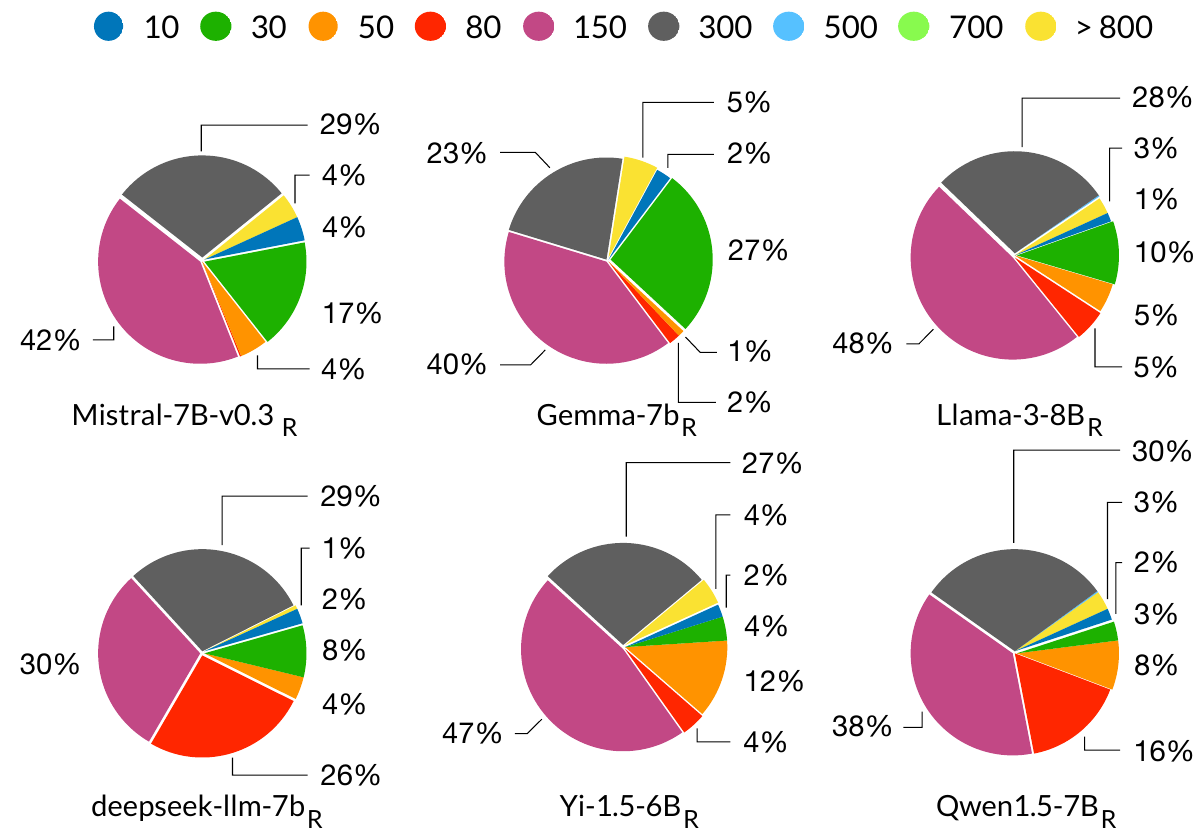}
		\caption{Distribution of \textit{MLTs} generated by \method{} in self-generated \textit{MLT} experiment. The models demonstrate a preference for generating responses with lengths of 150 and 300.}
		\label{fig:pie}
	\end{minipage}
	\hspace{0.03\textwidth} 
	\begin{minipage}[c]{0.4\textwidth}
		\centering
		\vspace{\fill}
		\scalebox{0.9}{
			\begin{tabular}{l|rc}
				\toprule
				\textbf{Model}                 & \textbf{FM} & \textbf{Avg WC} \\
				\midrule
				Mistral-7B-v0.3$_{\textsc{R}}$ & 73.40       & 279             \\
				gemma-7b$_{\textsc{R}}$        & 69.00       & 347             \\
				Llama-3-8B$_{\textsc{R}}$      & 88.40       & 215             \\
				deepseek-llm-7b$_{\textsc{R}}$ & 84.40       & 187             \\
				Yi-1.5-6B$_{\textsc{R}}$       & 81.40       & 236             \\
				Qwen1.5-7B$_{\textsc{R}}$      & 81.60       & 245             \\
				\bottomrule
			\end{tabular}
		}
		\captionof{table}{The FM score and average word count of \method{} with models in self-generated \textit{MLT} experiment. FM scores are notably high. Specifically, gemma-7b$_{\textsc{R}}$ recorded the lowest at 69.00, while Llama-3-8B$_{\textsc{R}}$ achieved the highest at 88.40.}
		\label{tab:FM arena no tl}
	\end{minipage}
\end{figure*}

\subsection{Do \textit{MLTs} actually influence the length of the generated content?}

To further investigate the effectiveness and scalability of \textit{MLTs}, we designed two additional experiments: multi \textit{MLT} generation experiment and self-generated \textit{MLT} experiment.

\begin{table*}[h]
	\centering
	\scalebox{0.75}{
		\begin{tabular}{lccccccc}
			\toprule
			\textbf{Model}       & \textbf{Type} & \textbf{ARC (chanllenge/easy)} & \textbf{HellaSwag} & \textbf{TruthfulQA} & \textbf{MMLU} & \textbf{Winogrande} & \textbf{GSM8K} \\
			\midrule
			Mistral-7B-v0.3      & vanilla       & 38.23/67.76                    & 48.57              & 46.02               & 34.94         & 62.04               & 26.46          \\
			-                    & \method       & 37.97/67.85                    & 47.83              & 47.12               & 37.88         & 62.83               & 27.52          \\
			\midrule
			gemma-7b             & vanilla       & 35.75/65.66                    & 45.95              & 41.13               & 32.44         & 57.14               & 23.58          \\
			-                    & \method       & 38.99/67.47                    & 45.40              & 45.65               & 31.67         & 60.30               & 25.93          \\
			\midrule
			Meta-Llama-3-8B      & vanilla       & 48.63/77.48                    & 58.89              & 51.41               & 50.91         & 71.74               & 44.96          \\
			-                    & \method       & 49.23/77.99                    & 59.12              & 51.90               & 50.16         & 71.19               & 46.63          \\
			\midrule
			deepseek-llm-7b-base & vanilla       & 50.94/79.92                    & 61.48              & 39.90               & 48.65         & 72.93               & 38.89          \\
			-                    & \method       & 51.37/79.55                    & 61.31              & 38.43               & 48.81         & 72.77               & 37.15          \\
			\midrule
			Yi-1.5-6B            & vanilla       & 51.62/79.25                    & 58.79              & 55.32               & 54.68         & 68.51               & 52.01          \\
			-                    & \method       & 51.28/79.46                    & 58.41              & 49.94               & 55.13         & 68.11               & 50.34          \\
			\midrule
			Qwen1.5-7B           & vanilla       & 46.67/77.53                    & 56.39              & 53.98               & 54.00         & 65.98               & 44.88          \\
			-                    & \method       & 47.27/76.68                    & 56.46              & 50.18               & 54.59         & 65.19               & 47.01          \\
			\bottomrule
		\end{tabular}
	}
	\vspace{-1mm}
	\caption{\label{tab:performace test}
		Comparison of the overall performance of six models with \method{} or vanilla, with scores computed on ARC, HellaSwag, TruthfulQA, MMLU, Winogrande and GSM8K. The overall performance of models using \method{} generally remains consistent with the base models with sft.
	}
\end{table*}

\paragraph{Multi \textit{MLT} Generation Experiment.}

To further validate the efficacy and robustness of \method, we assess its ability to control response length. We randomly sample 200 entries from Arena-Hard-Auto \citep{arenahard2024} and subject each to all target lengths (Table \ref{tab:target lengths}), culminating in 1,800 entries at last. Subsequently, we calculate the FM scores for each target length, using the original model as a baseline.

The results presented in Table \ref{tab:arena tl results} highlight the enhancements in model performance due to \method. The FM scores achieved by \method{} generally surpass those of the baseline models. Notably, even the well-performing Llama-3-8B$_\textsc{R}$ shows significant improvements. However, when the target length is 700, \method{} shows a decline in FM if the baseline model already achieves a certain score. In contrast, \method{} enhances performance if the baseline model is underperforming. This phenomenon is likely due to an imbalance in the $\mathcal{D}_{MLT}$, where responses of 700 words are infrequent and differ from the fine-tuning data of the baseline, potentially undermining performance. Overall, \method{} significantly improves model performance.

\paragraph{Self-generated \textit{MLT} Experiment.}

To validate \method{} in generating \textit{MLT} and responses under a non-\textit{TLG} scenario, we use the Arena-Hard-Auto dataset without providing \textit{MLTs}, thereby necessitating autonomous response generation by the model. We evaluate performance by cataloging the types and proportions of generated \textit{MLTs} (Figure \ref{fig:pie}) and evaluating response length using FM score at the target lengths corresponding to the \textit{MLTs} (Table \ref{tab:FM arena no tl}).

Models show a preference for producing responses with target lengths of 150 and 300. This inclination is likely attributable to the complex nature of the queries in the Arena-Hard-Auto, which require longer responses for problem resolution. In the non-\textit{TLG} scenario, the FM scores are notably high, with the Mistral-7B-v0.3$_{\mathbf{R}}$ recording the lowest at 73.40 and Llama-3-8B$_{\mathbf{R}}$ achieving the highest at 88.40. The word count across all models varies from 187 words to 347 words.

\subsection{Evaluation on Overall Performance}

To evaluate the impact of \method{} on overall performance, we conduct experiments utilizing six benchmark datasets: ARC \cite{clark2018think}, HellaSwag \citep{zellers2019hellaswag}, TruthfulQA \citep{lin-etal-2022-truthfulqa}, MMLU \citep{hendryckstest2021}, Winogrande \citep{sakaguchi2019winogrande} and GSM8K \citep{cobbe2021gsm8k}. These benchmarks provide a comprehensive assessment across different task types.
\texttt{lm-evaluation-harness} \citep{eval-harness} is employed to assess the overall performance.
Further details about the experiments on the experiment can be found in Appendix \ref{appendix:More Details of Other Tasks}.

Table \ref{tab:performace test} illustrates that \method{} marginally reduces performance on several tasks. Overall performance of models using Ruler generally remains consistent with the original models. The variations in scores are minimal, with changes within a very small range. Moreover, we observe that some models with Ruler actually show improvements in specific tasks. These improvements suggest that Ruler may contribute positively under certain conditions or in certain task types. This indicates that \method{} can significantly enhance the model's ability to follow length-based instructions without compromising its performance on the same data.

\section{Conclusion}

This study initially investigate the instruction following abilities of LLMs and introduces \textit{Target Length Generation Task (TLG)}. Additionally, we propose \method, a novel and model-angnostic method that controls generated length for LLMs. \method{} utilizes the \textit{MLT} and end-to-end training to enhance model performance. Experimental results demonstrate that substantial improvements in PM and FM scores across various models. Moreover, two additional experiments are conducted to further validate the efficacy of the proposed method. Finally, we assess overall performance  across six different benchmarks to demonstrate its superiority.

\section*{Limitations}

With the emergence of large language models (LLMs), an increasing number of applications are now utilizing LLMs. A particularly interesting aspect is the instruction-following capabilities of LLMs. In this paper, we analyze the capabilities of LLMs solely from the perspective of controlling generated length and propose a solution through \method. Instructions, which vary widely and represent a real-life scenario or application. We believe addressing the challenges or solving widespread issues across various instructions is crucial. We employ meta token to construct \method{} and argue that meta tokens offer more robust control over models than prompts do. Exploring how to develop and utilize models effectively with the help of tokens is a profoundly important question.

\section*{Ethical Statements}

This study concentrates on managing the output length of Large Language Models (LLMs). While our primary focus is on the length of generated content, we have not assessed the potential for producing toxic content. The research does not involve human participants, nor does it handle personal or sensitive information. We have used only open-source or suitably licensed resources, thereby complying with relevant standards. Additionally, all training data employed are open-source, ensuring the exclusion of any private or sensitive information.

\section*{Acknowledgements}
This work was supported by National Key Research and Development Program of China (2022YFF0902100), National Natural Science Foundation of China (Grant No. 62376262), the Natural Science Foundation of Guangdong Province of China (2024A1515030166), Shenzhen Science and Technology Innovation Program (KQTD20190929172835662), Shenzhen Basic Research Foundation (JCYJ20210324115614039).

\bibliography{anthology,custom}
\bibliographystyle{acl_natbib}

\clearpage

\onecolumn

\appendix
\section{Target Length Generation Task Deatils}
\label{appendix:Target Length Generation Task Deatils}

In this section, we present the experimental details of the \textit{Target Length Generation (TLG)}.

\subsection{\textit{TLG} Dataset}
\label{appendix:Dataset of TLG}
Dataset constructed for the \textit{TLG}, totaling 2,000 entries.

\begin{tcolorbox}[title=\textit{TLG} Dataset,colback=yellow!10!white,colframe=yellow!75!black]
	$\{$

	\hspace{0.5cm}\textbf{"id":}"0"

	\hspace{0.5cm}\textbf{"Instruction":}"How can I generate an AI model that can classify articles of clothing as shorts, skirts, or pants based on their descriptions?",

	\hspace{0.5cm}\textbf{"TargetLength":}"50"

	$\}$

	[...]

	$\{$

	\hspace{0.5cm}\textbf{"id":}"1999"

	\hspace{0.5cm}\textbf{"Instruction":}"You will be given several pieces of information about someone, and you will have to answer a question based on the information given.$\backslash$nJohn is taller than Bill. Mary is shorter than John. Question: Who is the tallest person?",

	\hspace{0.5cm}\textbf{"TargetLength":}"30"

	$\}$

\end{tcolorbox}

\subsection{Models \& Prompt Templates}
\label{appendix:Models & Prompt Templates}

In this appendix, we list the models in the \textit{TLG}, including their fullname, params, context length and vocab size. All models are downloaderd from Huggingface\footnote{\url{https://huggingface.co/}} and inference is executed using \texttt{vllm} \citep{kwon2023efficient}.

\begin{table*}[h]
	\centering
	\begin{tabular}{llrrr}
		\toprule
		\textbf{Model}                & \textbf{Model Full Name}  & \textbf{Params} & \textbf{Context Length} & \textbf{Vocab Size} \\
		\midrule
		Mistral                       & Mistral-7B-Instruct-v0.3  & 7B              & 32,768                  & 32,768              \\
		\midrule
		\multirow{2}{*}{Gemma}        & gemma-2b-it               & 2B              & 8,192                   & 256,000             \\
		                              & gemma-7b-it               & 7B              & 8,192                   & 256,000             \\
		\midrule
		\multirow{2}{*}{Llama3}       & Meta-Llama-3-8B-Instruct  & 8B              & 8,192                   & 128,256             \\
		                              & Meta-Llama-3-70B-Instruct & 70B             & 8,192                   & 128,256             \\
		\midrule
		\multirow{2}{*}{InternLM2}    & InternLM2-Chat-7B         & 7B              & 32,768                  & 92,544              \\
		                              & InternLM2-Chat-20B        & 20B             & 32,768                  & 92,544              \\
		\midrule
		\multirow{2}{*}{DeepSeek-LLM} & deepseek-llm-7b-chat      & 7B              & 4,096                   & 102,400             \\
		                              & deepseek-llm-67b-chat     & 67B             & 4,096                   & 102,400             \\
		\midrule
		\multirow{3}{*}{Yi-1.5}       & Yi-1.5-6B-Chat            & 6B              & 4,096                   & 64,000              \\
		                              & Yi-1.5-9B-Chat            & 9B              & 4,096                   & 64,000              \\
		                              & Yi-1.5-34B-Chat           & 34B             & 4,096                   & 64,000              \\
		\midrule
		\multirow{4}{*}{Qwen1.5}      & Qwen1.5-7B-Chat           & 7B              & 32,768                  & 151,936             \\
		                              & Qwen1.5-14B-Chat          & 14B             & 32,768                  & 151,936             \\
		                              & Qwen1.5-32B-Chat          & 32B             & 32,768                  & 151,936             \\
		                              & Qwen1.5-72B-Chat          & 72B             & 32,768                  & 151,936             \\
		\bottomrule
	\end{tabular}
	\caption{\label{tab:models}
		All models used in \textit{TLG} }
\end{table*}

\begin{table*}[h]
	\centering
	\scalebox{0.8}{
		\begin{tabular}{lll}
			\toprule
			\textbf{Model}                & \textbf{Prompt Template}                                                                   & \textbf{Eos Tokens}                                              \\
			\midrule
			Mistral                       & \texttt{<s>[INST] \{Instruction\} [/INST]}                                                 & \texttt{</s>}                                                    \\
			\midrule
			\multirow{2}{*}{Gemma}        & \texttt{<bos><start\_of\_turn>user\textbackslash n\{Instruction\}}                         & \multirow{2}{*}{\texttt{<eos>}}                                  \\
			                              & \texttt{<end\_of\_turn>\textbackslash n<start\_of\_turn>model\textbackslash n}             &                                                                  \\
			\midrule
			\multirow{3}{*}{Llama3}       & \texttt{<|begin\_of\_text|><|start\_header\_id|>user}                                      & \multirow{3}{*}{\texttt{<|end\_of\_text|>},\texttt{<|eot\_id|>}} \\
			                              & \texttt{<|end\_header\_id|>\textbackslash n\textbackslash n\{Instruction\}<|eot\_id|>}     &                                                                  \\
			                              & \texttt{<|start\_header\_id|>assistant<|end\_header\_id|>\textbackslash n\textbackslash n} &                                                                  \\

			\midrule
			\multirow{2}{*}{InternLM2}    & \texttt{<s><|im\_start|>user\textbackslash n\{Instruction\}}                               & \multirow{2}{*}{\texttt{</s>},     \texttt{<|im\_end|>}}         \\
			                              & \texttt{<|im\_end|>\textbackslash n<|im\_start|>assistant\textbackslash n}                 &                                                                  \\
			\midrule
			\multirow{2}{*}{DeepSeek-LLM} & \texttt{<|begin\_of\_sentence|>User: \{Instruction\}}                                      & \multirow{2}{*}{\texttt{<|end\_of\_sentence|>}}                  \\
			                              & \texttt{\textbackslash n\textbackslash nAssistant:}                                        &                                                                  \\
			\midrule
			\multirow{2}{*}{Yi-1.5}       & \texttt{<|im\_start|>user\textbackslash n\{Instruction\}<|im\_end|>}                       & \multirow{2}{*}{\texttt{<|im\_end|>},\texttt{<|endoftext|>}}     \\
			                              & \texttt{\textbackslash n<|im\_start|>assistant\textbackslash n}                            &                                                                  \\
			\midrule
			\multirow{3}{*}{Qwen1.5}      & \texttt{<|im\_start|>system\textbackslash nYou are a helpful assistant.}                   & \multirow{3}{*}{\texttt{<|im\_end|>},  \texttt{<|endoftext|>}}   \\
			                              & \texttt{<|im\_end|>\textbackslash n<|im\_start|>user\textbackslash n\{Instruction\}}       &                                                                  \\
			                              & \texttt{<|im\_end|>\textbackslash n<|im\_start|>assistant\textbackslash n}                 &                                                                  \\
			\bottomrule
		\end{tabular}
	}
	\caption{\label{tab:Models-templates}
		Prompt templates and Eos tokens for all models used in \textit{TLG}.
	}
\end{table*}

\subsection{Results on Different Target Length}
\label{appendix:Results on Different Target Length}

Here, we present the FM and PM scores of the models at all target lengths.

\subsubsection{\textit{Level:0}}

The PM and FM scores for each model at \textit{Level:0} are shown in Table \ref{tab:Closed-source Models Level:0 Results} and Table \ref{tab:Open-source Models Level:0 Results}.

\begin{table*}[!h]
	\centering
	\scalebox{0.9}{
		\begin{tabular}{lrrrrrrrrr}
			\toprule
			\multirow{3}{*}{\textbf{Model}} & \multirow{3}{*}{\textbf{Params}} & \multicolumn{8}{c}{\textit{Level:0}}                                                                                                                                                            \\
			\cmidrule(lr){3-10}
			                                &                                  & \multicolumn{2}{c}{10}               & \multicolumn{2}{c}{30} & \multicolumn{2}{c}{50} & \multicolumn{2}{c}{80}                                                                                 \\
			\cmidrule(lr){3-4}\cmidrule(lr){5-6}\cmidrule(lr){7-8}\cmidrule(lr){9-10}
			                                &                                  & PM                                   & FM                     & PM                     & FM                     & PM                & FM                & PM                & FM                \\
			\midrule
			Mistral                         & 7B                               & 30.73                                & 30.73                  & 18.60                  & 18.60                  & 16.87             & 16.87             & 15.45             & 28.64             \\
			\midrule
			\multirow{2}{*}{Gemma}          & 2B                               & 21.56                                & 21.56                  & 30.23                  & 30.23                  & 20.88             & 20.88             & 11.36             & 20.45             \\
			                                & 7B                               & 12.39                                & 12.39                  & 18.14                  & 18.14                  & 18.88             & 18.88             & 12.27             & 25.91             \\
			\midrule
			\multirow{2}{*}{Llama3}         & 8B                               & 45.41                                & 45.41                  & 35.35                  & 35.35                  & 33.73             & 33.73             & 24.09             & \underline{46.36} \\
			                                & 70B                              & \textbf{60.55}                       & \textbf{60.55}         & \textbf{66.05}         & \textbf{66.05}         & \textbf{61.45}    & \textbf{61.45}    & \textbf{46.82}    & \textbf{70.45}    \\
			\midrule
			\multirow{2}{*}{InternLM2}      & 7B                               & 17.89                                & 17.89                  & 6.98                   & 6.98                   & 1.20              & 1.20              & 1.36              & 3.64              \\
			                                & 20B                              & 20.64                                & 20.64                  & 8.84                   & 8.84                   & 2.81              & 2.81              & 4.55              & 8.18              \\
			\midrule
			\multirow{2}{*}{DeepSeek-LLM}   & 7B                               & \underline{58.26}                    & \underline{58.26}      & 25.12                  & 25.12                  & 17.67             & 17.67             & 13.18             & 26.36             \\
			                                & 67B                              & 46.79                                & 46.79                  & 20.47                  & 20.47                  & 22.09             & 22.09             & 19.09             & 32.73             \\
			\midrule
			\multirow{3}{*}{Yi-1.5}         & 6B                               & 39.91                                & 39.91                  & 23.72                  & 23.72                  & 20.08             & 20.08             & 10.91             & 20.45             \\
			                                & 9B                               & 47.71                                & 47.71                  & 23.72                  & 23.72                  & 17.27             & 17.27             & 13.64             & 29.55             \\
			                                & 34B                              & 45.41                                & 45.41                  & 27.44                  & 27.44                  & 20.48             & 20.48             & 23.18             & 42.73             \\
			\midrule
			\multirow{4}{*}{Qwen1.5}        & 7B                               & 31.19                                & 31.19                  & 25.58                  & 25.58                  & 22.89             & 22.89             & 17.73             & 30.45             \\
			                                & 14B                              & 45.87                                & 45.87                  & 28.84                  & 28.84                  & 26.51             & 26.51             & 12.27             & 25.45             \\
			                                & 32B                              & 46.79                                & 46.79                  & 33.95                  & 33.95                  & 29.32             & 29.32             & 20.91             & 35.91             \\
			                                & 72B                              & 39.45                                & 39.45                  & \underline{41.86}      & \underline{41.86}      & \underline{32.53} & \underline{32.53} & \underline{29.09} & 45.91             \\
			\bottomrule
		\end{tabular}
	}
	\caption{\label{tab:Open-source Models Level:0 Results}
		Results of open-source models of \textit{TLG} at \textit{Level:0}. The best-performing model in each target length is \textbf{in-bold}, and the second best is \underline{underlined}.
	}
\end{table*}

\begin{table*}[!h]
	\centering
	\scalebox{0.9}{
		\begin{tabular}{lrrrrrrrrr}
			\toprule
			\multirow{3}{*}{\textbf{Model}} & \multirow{3}{*}{\textbf{Params}} & \multicolumn{8}{c}{\textit{Level:0}}                                                                                                                                                            \\
			\cmidrule(lr){3-10}
			                                &                                  & \multicolumn{2}{c}{10}               & \multicolumn{2}{c}{30} & \multicolumn{2}{c}{50} & \multicolumn{2}{c}{80}                                                                                 \\
			\cmidrule(lr){3-4}\cmidrule(lr){5-6}\cmidrule(lr){7-8}\cmidrule(lr){9-10}
			                                &                                  & PM                                   & FM                     & PM                     & FM                     & PM                & FM                & PM                & FM                \\
			\midrule
			gpt-4-turbo                     & -                                & \textbf{89.45}                       & \textbf{89.45}         & \textbf{86.98}         & \textbf{86.98}         & \textbf{82.33}    & \textbf{82.33}    & \textbf{70.45}    & \underline{87.27} \\
			gpt-4o                          & -                                & \underline{83.49}                    & \underline{83.49}      & \underline{80.47}      & \underline{80.47}      & 71.08             & 71.08             & 61.82             & 82.27             \\
			gpt-3.5-turbo                   & -                                & 80.73                                & 80.73                  & 72.09                  & 72.09                  & 57.43             & 57.43             & 48.64             & 70.91             \\
			\midrule
			claude-3-haiku                  & -                                & 69.27                                & 69.27                  & 54.42                  & 54.42                  & 42.17             & 42.17             & 28.18             & 56.82             \\
			claude-3.5-sonnet               & -                                & 82.57                                & 82.57                  & 74.42                  & 74.42                  & \underline{75.50} & \underline{75.50} & \underline{68.18} & \textbf{92.27}    \\
			\bottomrule
		\end{tabular}
	}
	\caption{\label{tab:Closed-source Models Level:0 Results}
		Results of closed-source models of \textit{TLG} at \textit{Level:0}. The best-performing model in each target length is \textbf{in-bold}, and the second best is \underline{underlined}.
	}
\end{table*}

\subsubsection{\textit{Level:1}}

The PM and FM scores for each model at \textit{Level:1} are shown in Table \ref{tab:Open-source Models Level:1 Results} and Table \ref{tab:Closed-source Models Level:1 Results}.

\begin{table*}[!h]
	\centering
	\scalebox{0.9}{
		\begin{tabular}{lrrrrrrr}
			\toprule
			\multirow{3}{*}{\textbf{Model}} & \multirow{3}{*}{\textbf{Params}} & \multicolumn{6}{c}{\textit{Level:1}}                                                                                                                 \\
			\cmidrule(lr){3-8}
			                                &                                  & \multicolumn{2}{c}{150}              & \multicolumn{2}{c}{300} & \multicolumn{2}{c}{500}                                                             \\
			\cmidrule(lr){3-4}\cmidrule(lr){5-6}\cmidrule(lr){7-8}
			                                &                                  & PM                                   & FM                      & PM                      & FM                & PM                & FM                \\
			\midrule
			Mistral                         & 7B                               & 17.86                                & 41.84                   & 14.77                   & 70.04             & 17.94             & 30.94             \\
			\midrule
			\multirow{2}{*}{Gemma}          & 2B                               & 17.35                                & 32.65                   & 7.17                    & 33.33             & 2.69              & 7.17              \\
			                                & 7B                               & 18.88                                & 42.35                   & 12.24                   & 51.90             & 4.93              & 13.00             \\
			\midrule
			\multirow{2}{*}{Llama3}         & 8B                               & \underline{38.27}                    & \underline{70.92}       & \textbf{27.00}          & \underline{78.90} & 25.11             & 47.09             \\
			                                & 70B                              & \textbf{55.10}                       & \textbf{85.71}          & 22.36                   & \textbf{88.61}    & \underline{35.43} & \textbf{59.64}    \\
			\midrule
			\multirow{2}{*}{InternLM2}      & 7B                               & 9.18                                 & 20.92                   & 5.91                    & 37.55             & 11.21             & 22.42             \\
			                                & 20B                              & 9.69                                 & 22.96                   & 9.28                    & 45.99             & 13.90             & 32.29             \\
			\midrule
			\multirow{2}{*}{DeepSeek-LLM}   & 7B                               & 15.31                                & 37.24                   & 18.14                   & 60.76             & 19.28             & 33.18             \\
			                                & 67B                              & 9.18                                 & 34.69                   & 19.83                   & 71.73             & 21.08             & 39.01             \\
			\midrule
			\multirow{3}{*}{Yi-1.5}         & 6B                               & 18.88                                & 46.94                   & 12.66                   & 62.45             & 18.39             & 35.87             \\
			                                & 9B                               & 12.76                                & 33.16                   & 12.66                   & 53.59             & 26.46             & 44.39             \\
			                                & 34B                              & 25.51                                & 58.67                   & \underline{24.05}       & 78.48             & 28.70             & \underline{57.40} \\
			\midrule
			\multirow{4}{*}{Qwen1.5}        & 7B                               & 9.69                                 & 29.59                   & 7.17                    & 61.60             & 26.01             & 44.39             \\
			                                & 14B                              & 5.61                                 & 16.84                   & 10.97                   & 56.12             & \textbf{37.67}    & 54.71             \\
			                                & 32B                              & 20.92                                & 43.37                   & 14.77                   & 53.59             & 31.39             & 50.22             \\
			                                & 72B                              & 13.27                                & 35.20                   & 12.66                   & 64.98             & 28.70             & 46.19             \\
			\bottomrule
		\end{tabular}
	}
	\caption{\label{tab:Open-source Models Level:1 Results}
		Results of open-source models of \textit{TLG} at \textit{Level:1}. The best-performing model in each target length is \textbf{in-bold}, and the second best is \underline{underlined}.
	}
\end{table*}

\begin{table*}[!h]
	\centering
	\scalebox{0.9}{
		\begin{tabular}{lrrrrrrr}
			\toprule
			\multirow{3}{*}{\textbf{Model}} & \multirow{3}{*}{\textbf{Params}} & \multicolumn{6}{c}{\textit{Level:1}}                                                                                                                 \\
			\cmidrule(lr){3-8}
			                                &                                  & \multicolumn{2}{c}{150}              & \multicolumn{2}{c}{300} & \multicolumn{2}{c}{500}                                                             \\
			\cmidrule(lr){3-4}\cmidrule(lr){5-6}\cmidrule(lr){7-8}
			                                &                                  & PM                                   & FM                      & PM                      & FM                & PM                & FM                \\
			\midrule
			gpt-4-turbo                     & -                                & \underline{68.37}                    & \underline{93.88}       & \underline{22.36}       & \underline{83.97} & \textbf{52.91}    & \textbf{78.48}    \\
			gpt-4o                          & -                                & 60.20                                & 88.78                   & 15.61                   & 71.31             & 25.56             & 50.22             \\
			gpt-3.5-turbo                   & -                                & 54.08                                & 79.59                   & 15.19                   & 81.86             & 39.46             & 65.92             \\
			\midrule
			claude-3-haiku                  & -                                & 43.88                                & 76.02                   & 19.41                   & 75.95             & \underline{44.84} & \underline{69.51} \\
			claude-3.5-sonnet               & -                                & \textbf{76.53}                       & \textbf{97.45}          & \textbf{24.05}          & \textbf{88.61}    & 31.84             & 64.57             \\
			\bottomrule
		\end{tabular}
	}
	\caption{\label{tab:Closed-source Models Level:1 Results}
		Results of closed-source models of \textit{TLG} at \textit{Level:1}. The best-performing model in each target length is \textbf{in-bold}, and the second best is \underline{underlined}.
	}
\end{table*}

\subsubsection{\textit{Level:2}}

The PM and FM scores for each model at \textit{Level:2} are shown in Table \ref{tab:Open-source Models Level:2 Results} and Table \ref{tab:Closed-source Models Level:2 Results}.

\begin{table*}[!h]
	\centering
	\scalebox{0.9}{
		\begin{tabular}{lrrrrr}
			\toprule
			\multirow{3}{*}{\textbf{Model}} & \multirow{3}{*}{\textbf{Params}} & \multicolumn{4}{c}{\textit{Level:2}}                                                                    \\
			\cmidrule(lr){3-6}
			                                &                                  & \multicolumn{2}{c}{700}              & \multicolumn{2}{c}{>800}                                         \\
			\cmidrule(lr){3-4}\cmidrule(lr){5-6}
			                                &                                  & PM                                   & FM                       & PM                & FM                \\
			\midrule
			Mistral                         & 7B                               & 3.04                                 & 6.96                     & 4.25              & 4.25              \\
			\midrule
			\multirow{2}{*}{Gemma}          & 2B                               & 0.00                                 & 0.00                     & 0.47              & 0.47              \\
			                                & 7B                               & 0.87                                 & 0.87                     & 0.00              & 0.00              \\
			\midrule
			\multirow{2}{*}{Llama3}         & 8B                               & 16.09                                & 21.74                    & 20.28             & 20.28             \\
			                                & 70B                              & \underline{24.35}                    & \textbf{33.48}           & \textbf{49.53}    & \textbf{49.53}    \\
			\midrule
			\multirow{2}{*}{InternLM2}      & 7B                               & 18.70                                & 23.91                    & 20.75             & 20.75             \\
			                                & 20B                              & 17.39                                & 22.61                    & 17.45             & 17.45             \\
			\midrule
			\multirow{2}{*}{DeepSeek-LLM}   & 7B                               & 9.13                                 & 13.48                    & 12.74             & 12.74             \\
			                                & 67B                              & 9.13                                 & 13.91                    & 9.91              & 9.91              \\
			\midrule
			\multirow{3}{*}{Yi-1.5}         & 6B                               & 12.61                                & 16.96                    & 24.06             & 24.06             \\
			                                & 9B                               & 22.17                                & \underline{31.74}        & \underline{26.89} & \underline{26.89} \\
			                                & 34B                              & 22.17                                & 30.87                    & 20.28             & 20.28             \\
			\midrule
			\multirow{4}{*}{Qwen1.5}        & 7B                               & 12.17                                & 17.83                    & 5.66              & 5.66              \\
			                                & 14B                              & 15.22                                & 21.30                    & 6.60              & 6.60              \\
			                                & 32B                              & \textbf{23.91}                       & 31.30                    & 18.87             & 18.87             \\
			                                & 72B                              & 6.09                                 & 10.43                    & 1.42              & 1.42              \\
			\bottomrule
		\end{tabular}
	}
	\caption{\label{tab:Open-source Models Level:2 Results}
		Results of open-source models of \textit{TLG} at \textit{Level:2}. The best-performing model in each target length is \textbf{in-bold}, and the second best is \underline{underlined}.
	}
\end{table*}

\begin{table*}[!h]
	\centering
	\scalebox{0.9}{
		\begin{tabular}{lrrrrr}
			\toprule
			\multirow{3}{*}{\textbf{Model}} & \multirow{3}{*}{\textbf{Params}} & \multicolumn{4}{c}{\textit{Level:2}}                                                                    \\
			\cmidrule(lr){3-6}
			                                &                                  & \multicolumn{2}{c}{700}              & \multicolumn{2}{c}{>800}                                         \\
			\cmidrule(lr){3-4}\cmidrule(lr){5-6}
			                                &                                  & PM                                   & FM                       & PM                & FM                \\
			\midrule
			gpt-4-turbo                     & -                                & \textbf{49.57}                       & \underline{62.61}        & 31.13             & 31.13             \\
			gpt-4o                          & -                                & \underline{46.09}                    & \textbf{64.78}           & \underline{79.72} & \underline{79.72} \\
			gpt-3.5-turbo                   & -                                & 35.65                                & 50.43                    & 41.04             & 41.04             \\
			\midrule
			claude-3-haiku                  & -                                & 39.57                                & 51.74                    & 49.06             & 49.06             \\
			claude-3.5-sonnet               & -                                & 36.52                                & 53.04                    & \textbf{91.04}    & \textbf{91.04}    \\
			\bottomrule
		\end{tabular}
	}
	\caption{\label{tab:Closed-source Models Level:2 Results}
		Results of closed-source models of \textit{TLG} at \textit{Level:2}. The best-performing model in each target length is \textbf{in-bold}, and the second best is \underline{underlined}.
	}
\end{table*}

\section{$\mathcal{D}_{MLT}$ Data Creation}
\label{appendix:Data Creation}

\begin{algorithm}[h]
	\caption{$\mathcal{D}_{MLT}$ Data Creation}
	\label{algorithm1}
	\begin{algorithmic}[1]
		\REQUIRE Word count function $L(\cdot)$, meta length tokens $MLTs=\{MLT_0,MLT_1,\cdots\}$
		\renewcommand{\algorithmicrequire}{ \textbf{Input:}}
		\renewcommand{\algorithmicensure}{\textbf{Output:}} %
		\REQUIRE Initial dataset $\mathcal{D}$
		\ENSURE $\mathcal{D}_{MLT}$
		\STATE $\mathcal{D}_{MLT} \leftarrow \{\}$
		\FOR{each tuple $(x, y)$ in $D$}
		\STATE $mlt \leftarrow$ None
		\FOR{each $MLT$ in $MLTs$}
		\IF{$L(y)>\mathrm{lb}_{MLT}$ \AND $L(y)\leq\mathrm{ub}_{MLT}$}
		\STATE $mlt \leftarrow MLT$
		\STATE \textbf{break}
		\ENDIF
		\ENDFOR
		\IF{$mlt$ is not None}
		\STATE $\mathcal{D}_{MLT} \leftarrow \mathcal{D}_{MLT} \cup \{(x, mlt, y)\}$
		\ENDIF
		\ENDFOR
		\RETURN $\mathcal{D}_{MLT}$
	\end{algorithmic}
\end{algorithm}

\section{Experiments Details}
\label{appendix:Experiments Details}

\subsection{\textit{MLT} in Datasets}
\label{appendix:MLT in Datasets}

To obtain data with varying response lengths for composing $\mathcal{D}_{MLT}$, particularly those responses exceeding 500, we integrateg data from OpenHermes2.5 \citep{OpenHermes2.5}, LongForm \citep{koksal2023longform} and ELI5 \citep{fan2019eli5}. We calculate the word count for each response in every dataset, allowing us to statistically analyze the \textit{MLT} distribution, shown in Table \ref{tab:MLT in Datasets}.

\begin{table*}[h]
	\centering
	\begin{tabular}{lccc}
		\toprule
		\multirow{2}{*}{\textbf{\textit{MLT}}} & \textbf{OpenHermes2.5} & \textbf{LongForm}          & \textbf{ELI5}

		\\
		                                       & \citep{OpenHermes2.5}  & \citep{koksal2023longform} & \citep{fan2019eli5} \\
		\midrule
		\texttt{[MLT:10]}                      & 28,552                 & 586                        & 3,280               \\
		\texttt{[MLT:30]}                      & 16,860                 & 1,428                      & 14,143              \\
		\texttt{[MLT:50]}                      & 18,867                 & 1,236                      & 17,597              \\
		\texttt{[MLT:80]}                      & 18,014                 & 852                        & 15,926              \\
		\midrule
		\texttt{[MLT:150]}                     & 37,515                 & 1,037                      & 19,103              \\
		\texttt{[MLT:300]}                     & 7,526                  & 252                        & 2,555               \\
		\texttt{[MLT:500]}                     & 1,495                  & 140                        & 682                 \\
		\midrule
		\texttt{[MLT:700]}                     & 193                    & 101                        & 203                 \\
		\texttt{[MLT:800]}                     & 1,809                  & 2,465                      & 3,808               \\
		\bottomrule
	\end{tabular}
	\caption{\label{tab:MLT in Datasets}
		\textit{MLT} distribution in each dataset. The OpenHermes2.5 excludes the data utilized in \textit{TLG}. The LongForm and ELI5 employs its training, validation, and test sets simultaneously. When multiple answers are available in the dataset, the longest answer is selected as the final response.
	}
\end{table*}

\subsection{More Details of Training}
\label{appendix:More Details of Training}

\paragraph{More details of training.}

We use 4*A100 with 80GB Nvidia GPUs to train the models. The training utilizes both bf16 and tensor tf32 precision formats. The per-device training batch size is set to 4, with gradient accumulation is 8 steps. A cosine learning rate scheduler is applied, starting with an initial learning rate of 2e-5 and a warmup ratio of 0.05. All models are trained for 3 epochs. Additionally, log is set to print every 5 steps.

\paragraph{Loss.}

\begin{figure}[H]
	\centering
	\includegraphics[width=\textwidth]{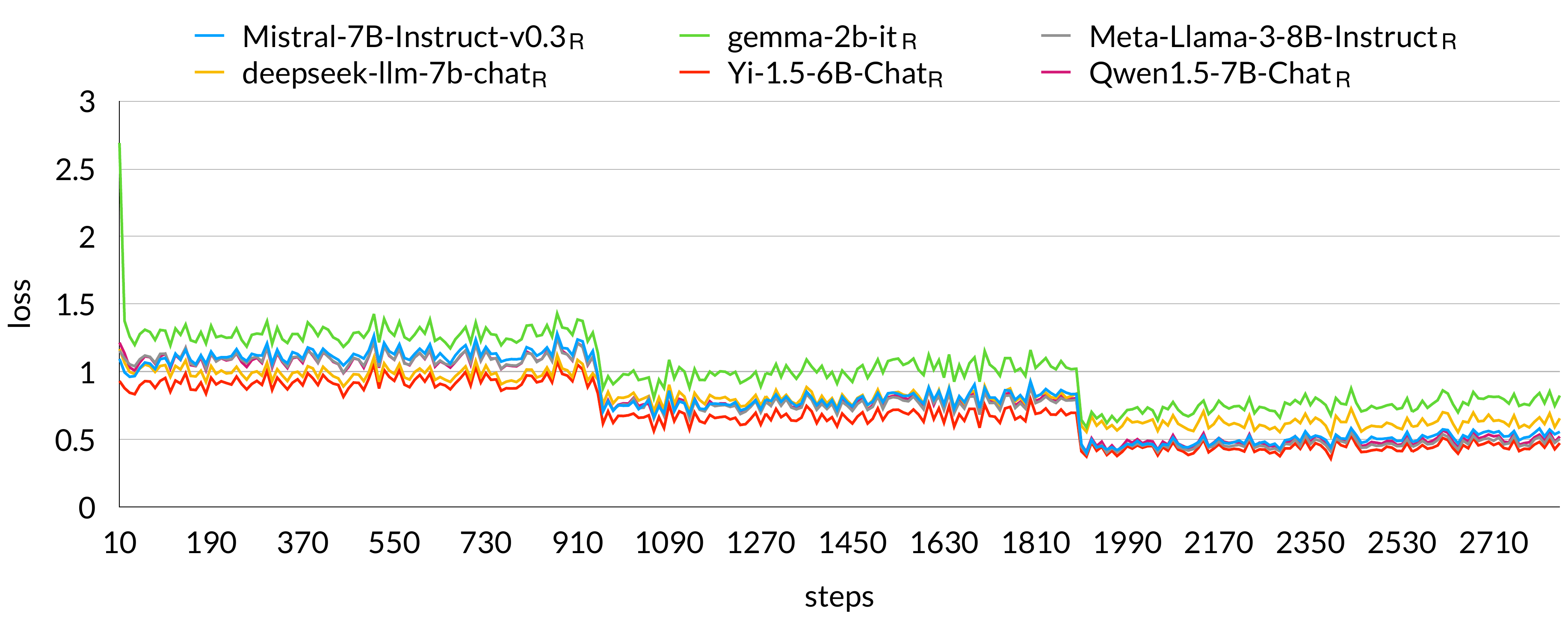}
	\caption{Training loss for models.}
	\label{fig:loss}
\end{figure}

We document the changes in training loss for all models, as shown in Figure \ref{fig:loss}.

\subsection{Multi \textit{MLT} generation experiment}
\label{appendix:Multi MLT generation experiment}

Here is the results in multi \textit{MLT} generation experiment.

\subsection{More Details of Other Tasks}
\label{appendix:More Details of Other Tasks}

We tested the \method on six  benchmarks (ARC \cite{clark2018think}, HellaSwag \citep{zellers2019hellaswag}, TruthfulQA \citep{lin-etal-2022-truthfulqa}, MMLU \citep{hendryckstest2021}, Winogrande \citep{sakaguchi2019winogrande} and GSM8K \citep{cobbe2021gsm8k}) to examine whether the performance of the fine-tuned models varies on different tasks. We employ 25-shot in ARC, 10-shot setting in Hellaswag, 5-shot setting in MMLU, 0-shot setting in TruthfulQA, 5-shot setting in Winogrande and 5-shot in GSM8K.

\end{document}